\pdfoutput=1

\documentclass[11pt]{article}

\usepackage[]{acl}

\usepackage{times}
\usepackage{latexsym}

\usepackage[T1]{fontenc}

\usepackage[utf8]{inputenc}

\usepackage{microtype}

\usepackage{multirow}
\usepackage{bm}
\usepackage{xspace}
\usepackage{balance}
\usepackage{amsmath}
\usepackage{graphicx}
\usepackage{url}
\usepackage{color,xcolor}
\usepackage{enumitem}

\newcommand{\eat}[1]{}
\newcommand{\paratitle}[1]{\vspace{1ex}\noindent \textbf{#1}}
\let\oldhat\hat
\renewcommand{\vec}[1]{\bm{#1}}
\renewcommand{\hat}[1]{\vec{\oldhat{#1}}}

\newcommand{\ie}{\emph{i.e., }\xspace}

\newcommand{\aka}{\emph{a.k.a., }\xspace}

\newcommand{\argmax}{\mathop{\arg\max}}

\title{Incorporating Causal Analysis into Diversified and Logical Response Generation}

\author{Jiayi Liu\textsuperscript{1}, Wei Wei\textsuperscript{2,3}\thanks{\textsuperscript{*}Corresponding author.}, Zhixuan Chu\textsuperscript{4}, Xing Gao\textsuperscript{1}, Ji Zhang\textsuperscript{1}, Tan Yan\textsuperscript{4}, Yulin Kang\textsuperscript{4} \\
  \textsuperscript{1}Alibaba Group, China \\
   \textsuperscript{2}CCIIP Laboratory, Huazhong University of Science and Technology, China \\  
   \textsuperscript{3}Joint Laboratory of HUST and Pingan Property \& Casualty Research (HPL), China\\
   \textsuperscript{4}Ant Group, China \\
  {\tt \{ljy269999,chuzhixuan.czx,gaoxing.gx,zj122146\}@alibaba-inc.com} \\
  {\tt weiw@hust.edu.cn}, 
  {\tt tanyan.ty@gmail.com},
  {\tt yulin.kyl@antgroup.com} \\ 
}

\begin{document}
\maketitle
\vspace{-8mm}
\begin{abstract}
\vspace{-2mm}
Although the Conditional Variational Auto-Encoder (CVAE) model can generate more diversified responses than the traditional Seq2Seq model, the responses often have low relevance with the input words or are illogical with the question. A causal analysis is carried out to study the reasons behind, and a methodology of searching for the mediators and mitigating the confounding bias in dialogues is provided. Specifically, we propose to predict the mediators to preserve relevant information and auto-regressively incorporate the mediators into generating process. Besides, a dynamic topic graph guided conditional variational auto-encoder (TGG-CVAE) model is utilized to complement the semantic space and reduce the confounding bias in responses. Extensive experiments demonstrate that the proposed model is able to generate both relevant and informative responses, and outperforms the state-of-the-art in terms of automatic metrics and human evaluations.
\end{abstract}

\section{Introduction}\label{sec:intro}
\vspace{-2mm}

With recent advances in deep learning and readily available large-scale dialogue data, generation-based methods have become one of the most prevailing methods for building dialogue systems. Based on the Seq2seq framework \cite{sutskever2014sequence,cho2014learning}, generation-based models learn to map the input post to its corresponding response through an encoding-decoding strategy and are trained in end-to-end manners \cite{shang2015neural,sordoni2015neural,vinyals2015neural}. However, Seq2seq model tends to produce generic and safe responses \cite{li2015diversity} such as ``So am I'' or ``I don’t know''. Researchers conjecture that the cause of this phenomenon is that one certain post can be replied by multiple responses (\ie one-to-many mapping), and the maximum likelihood estimation (MLE) training would average out these responses and produce a more bland and generic candidate.

\begin{table}
\centering
\begin{tabular}{ll}
\hline
Post:  & Have you had dinner? \\
\hline
Response1: & \textcolor{red}{Yeah, sure!}  \\
Response2: & \textcolor{red}{Yes,} I had it at McDonald's. \\
Response3: & \textcolor{red}{Nope,} I'm busy with my work.  \\
Response4: & \textcolor{red}{Yes, I've had it.} I tried a nearby \\  
           & restaurant that  features Thai food. \\
\hline
\end{tabular}
\label{tab1}
\caption{An illustration of a general question and its multiple valid responses. The {\it direct responding semantics} (marked in red) are semantically homogeneous because they have to reply the issue directly. The {\it supplementary semantics} are more diversified because they add more information to explain or supplement the corresponding {\it direct responding semantics}.}
\vspace{-0.5cm}
\end{table}

To tackle this problem and model the one-to-many mapping relationships in dialogues, \cite{zhao2017learning} firstly leverages Conditional Variational Auto-Encoder (CVAE) model to map the input post into a semantic distribution, instead of a fixed vector as used in the vanilla Seq2seq model. The decoder then decodes the sampled points from the semantic distribution to generate corresponding responses. This model significantly increases the diversity of responses, but it is hard to train as the valid responses are too few to shape a clear semantic distribution for each post. As a result, the CVAE model is inclined to learn some spurious statistical cues for predicting diversified words, which may have very low relevance with the input post. Other studies focus on re-using the model's components to fit the multiplicity of dialogues, for instance, the multiple mechanisms used in \cite{zhou2017mechanism} and \cite{zhou2018elastic}, the multi-head attention used in  \cite{tao2018get,Liu2022DepthAwareAS}, and reinforced methods\cite{Qiu2021ReinforcedHB}. The most-related work in this line is the Multi-Mapping and Posterior Mapping Selection (MMPMS) \cite{chen2019generating} model, which directly builds multiple mapping modules to learn diversified semantics and generate responses. However, these studies haven’t considered the intrinsic nature of this one-to-many phenomenon in dialogues.

We always face the trade-off between the accuracy of response and diversity of semantics, and cannot directly generate relevant and diversified responses from the original input post. To solve this dilemma and examine the nature of dialogues, we introduce the causal inference analysis \cite{pearl1995causal,pearl2000causality} into the dialogue generation task. Here, we assume between the input post and outcome response, there exists one mediator. The mediator can easily capture the relevant but simple response from the input post ($\text{input post} \rightarrow \text{mediator}$) and also can pass the learned information to the outcome so as to preserve the relevance. In addition, when generating the diversified responses, the sampling steps in prior and posterior distributions of CVAE will act as the confounders between the input and outcome response. Therefore, we establish one causal graph including the mediator, confounder, input post, and segmented responses, i.e, {\it direct responding semantics} and {\it supplementary semantics}, to facilitate the information transmission and enrichment, and preserve the relevance and logicality.

Based on the above causal analysis, this work presents a unified end-to-end sentence-level auto-regressive model (SLARM) to predict the mediator and mitigate the confounding bias in generating diverse responses. We concrete the mediator by predicting the direct responding semantics, and leverage this mediator in an auto-regressive manner for response generation. A dialogue topic graph enhanced CVAE model with a larger semantic space is proposed to reduce the confounding bias in CVAE model, and thus make sure the transition is smooth and natural. In conclusion, the contributions of this work are three-fold: 

\begin{enumerate}
    \item It provides an in-depth analysis of the underlying causality involved in the dialogue generation task, and proposed a methodology of searching for the mediators and mitigating the confounding bias in dialogues.  
    \item It proposes an innovative dialogue generation model based on the established causal graph with mediator and confounder. The model predicts the {\it direct responding semantics} as mediators and generate the {\it supplementary semantics} in a unified auto-regressive manner using the proposed TGG-CVAE part to mitigate the confounding bias.
    \item It conducts broad experiments on a real-world dialogue dataset, which demonstrates that our proposed approach outperforms the state-of-the-art methods and has the capability of enhancing the diversity of responses without the sacrifice of relevance.
\end{enumerate}

\section{Related Works}
\vspace{-2mm}
\label{sec:r_work}
\noindent\textbf{Diversified Generation models.}
Some researchers suggest that the maximum-likelihood training objective used in the seq2seq model will average out the targets and result in safe and commonplace responses. Several attempts have been made to tackle this problem by proposing diversity-promoting objective functions, such as Maximum Mutual Information (MMI) \cite{li2015diversity}, Inverse Token Frequency Loss (ITF) \cite{nakamura2018another}. Although these studies help mitigate the safe response problem, their performance is far from satisfactory. Recently, researchers have discovered that incorporating additional information can lead to more diverse responses. Such methods include predicting keywords to guide the generation process \cite{mou2016sequence,yao2017towards}, and using latent variables such as \cite{zhao2017learning,gao2019generating,gao2019discrete,Wei2019EmotionawareCM,Wei2021TargetguidedEC}. Some recent studies focus on the one-to-many relationship between a certain post and its multiple valid responses, which is a common phenomenon in real dialogues. For instance, \cite{zhou2017mechanism} and \cite{zhou2018elastic} model the one-to-many mapping relationships through multiple latent mechanisms and leverage diverse mechanisms to enhance the diversity of generated responses. \cite{tao2018get} leverages the multi-head attention to focus on different parts of the input post and generate diverse responses. The state-of-the-art model in this line is the Multi-Mapping and Posterior Mapping Selection (MMPMS) \cite{chen2019generating} model, which directly builds multiple mapping modules to learn diversified semantics and generate responses.

\noindent\textbf{Causal Inference.}
Causal inference \citep{pearl2000causality,rubin2005causal} has been an attractive research topic for a long time since it provides an effective way to uncover causal relationships in real-world problems. Nowadays, the combination of the incisive ideas in the causal inference and various deep learning model can help improve existing methodologies in a wide range of fields, such as treatment effect estimation with observational data \citep{li2017matching,chu2020matching,chu2022learning}, causality analysis of graph networked data \citep{chu2021graph}, continual learning \citep{hu2021distilling, chu2020continual}, natural language processing task \citep{yang2021causal,niu2021counterfactual,abbasnejad2020counterfactual}, few-shot learning \citep{yue2020interventional, yue2021counterfactual}, domain adaptation~\citep{bengio2019meta}, clinical trials \citep{chu2022multi}, finance \citep{atanasov2016shock}, accounting \citep{gow2016causal}, marketing campaigns \citep{chu2022hierarchical} and so on. It is very challenging to choose or define proper confounders and mediators so as to construct one reasonable causal graph for different new tasks. A confounder is related to both cause and effect in a study, and a mediator explains the process by which cause and effect are related.  In this work, we aim to incorporate causal inference into the dialogue generation task to help the model balance the relevance and diversity of response semantics. 

\section{Causal Analysis}
\vspace{-2mm}
\label{sec:causal}

\begin{figure}[th!]
\centering
\vspace{-3mm}
\includegraphics[scale=0.1]{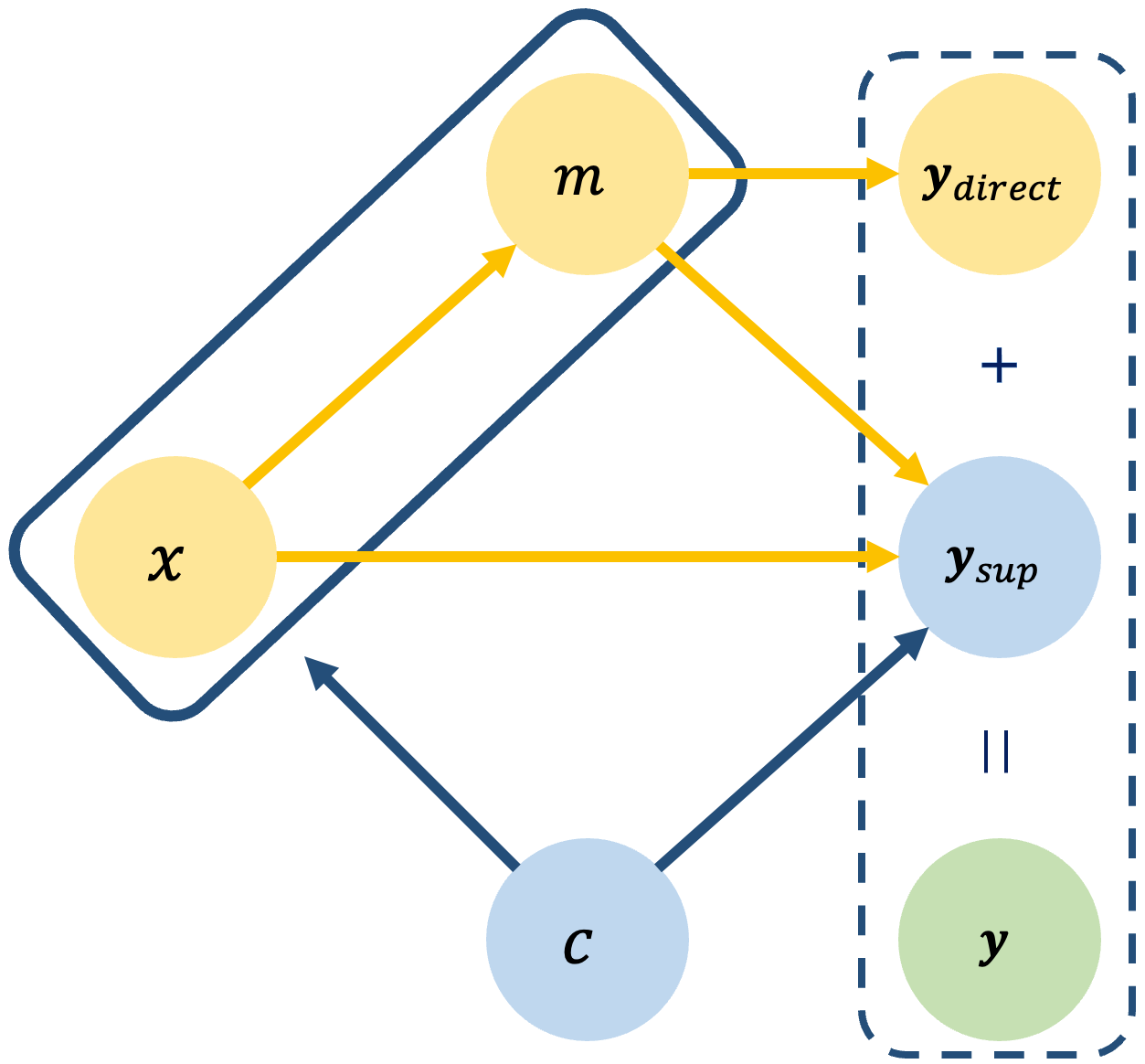}
\caption{The causal graph of dialogue generation task. The dialogue generation task can be naturally abstracted to one causal graph involving input post $x$,  confounder $c$, mediators $m$, direct responding semantics $\vec{y}_{direct}$, and supplementary semantics $\vec{y}_{sup}$. The direct responding semantics $\vec{y}_{direct}$ is the proxy variable of mediator $m$. The complete response $y$ consists of direct responding semantics $\vec{y}_{direct}$, and supplementary semantics $\vec{y}_{sup}$.}
\label{graph}
\vspace{-0.3cm}
\end{figure}


In this section, we introduce the causal inference analysis \cite{pearl1995causal,pearl2000causality,yao2021survey} into this task and define the mediator and confounder in the dialogue generation causal graph. A mediator is determined by input post and has causal effects on outcome response, and a confounder has causal effects on both input post and outcome response. Our objective is to leverage the causal relationship involved in the established causal graph to increase the diversity of response semantics, but at the same time, not to reduce the relevance of response to input post.

We assume there exists one mediator between the input post and outcome response. The mediator can easily capture the relevant but simple response from the input post ($\text{input post} \rightarrow \text{mediator}$) and also can pass the learned information to outcome so as to preserve the relevance ($\text{mediator} \rightarrow \text{outcome response}$). Except for the path via mediator, the input post is also directly predictive of the outcome response ($\text{input post} \rightarrow \text{outcome response}$). In addition, we propose to use the CVAE to increase the diversity of responses. However, the sampling steps in prior and posterior distributions of CVAE will act as the confounder between the input combination (input post and mediator) and outcome response ($\text{input combination} \leftarrow \text{confounder} \rightarrow \text{outcome response}$). Because the input combination and response pairs maybe do not conform to the assumed prior or posterior distributions of CVAE, this confounding bias may make the model learn the spurious statistical cues for the prediction of diversified response, resulting in some linguistically similar but inconsistent or irrelevant expressions in the generated sentences. Therefore, reducing the confounding bias is essential for the dialogue generation task.

Corresponding to the above causal relationship, we split the complete response into two parts, i.e., {\it direct responding semantics} and {\it supplementary semantics}, as shown in Figure \ref{graph}. The {\it direct responding semantics} represents the semantic part that can be directly leveraged to answer the input question. The direct responding semantics is the proxy variable of the mediator. The {\it Supplementary semantics} represents the peripheral semantic part that is either an explanation, a supplement, or an extension of the {\it direct responding semantics}. The {\it direct responding semantics} is semantically homogeneous because it has to solve the issue directly, and the {\it supplementary semantics} is more diversified because it adds more information to explain or supplement the {\it direct responding semantics}, or even change the topics to make conversation continue. 


Although the high-quality observations of the mediators can reduce the confounding bias hidden in the causal structure by reducing the possibility of counting on the confounders, it is not enough to attain one relevant and diversified response in the complex dialogue generation task. In addition, unlike the mediator that can be represented by direct responding semantics, it is very challenging to define and construct the exact confounders clearly. Therefore, due to the complex causal graph and hidden confounders, the front-door and back-door adjustments \cite{glymour2016causal, pearl2018book} for reducing the confounding bias cannot be easily applied. Therefore, instead of the conventional causal intervention based on Pearl's do-calculus \cite{pearl2018book}, we propose to exploit the dialogue topic graph to complement the semantic space and assign more relevant information into CVAE, which can enhance the diversity and keep the relevance of input post simultaneously.

\section{Proposed Model}
\vspace{-2mm}
\label{sec:model}

Our response generation task is defined as follows. Given an input post $\vec{x}= \{x_1,x_2,\cdots, x_{T}\}$, the problem is to generate the corresponding response sequence $\vec{y} = \{y_1,y_2,\cdots, y_{T^{''}}\}$, where $T$ is the length of the post and $T^{''}$ is the length of response.

To address this problem, we propose to generate the response sequence in a sentence-level auto-regressive manner. Firstly, we predict the proxy variable of the mediator by maximizing the log-likelihood of the following formula:
\begin{equation}
  \vec{y}^{*}_{direct} = \argmax\Pr(\vec{y}|\vec{x}).
\end{equation}
As mentioned above, this process produce general responses, but they are closely related to the input post and may help determine where the conversation should go. Hence, we preserve the causal path (input post → mediator) and then we can transmit the learned information to outcome so as to preserve the relevance 
(mediator → outcome response).

Then, an Sentence Level Auto-Regressive generating Model (SLARM) is proposed to produce diverse and informative responses based on the mediator and the dialogue topic graph. We first propose to utilize the predicted mediator in an auto-regressive manner:
\begin{equation}
  \vec{y}^{*}_{sup} = \argmax\Pr(\vec{y}|\vec{x},\vec{y}^{*}_{direct}),
\end{equation}
and then build a topic graph enhanced CVAE model to mitigate the confounding bias in traditional CVAE models. The auto-regressive training manner serves like a prompt to naturally inject the mediator into generation process, and the topic graph provides dynamic guidance to prevent the CVAE model from off-the-topic deviation and complement the semantic space.

\subsection{Mediator Predictor}

As aforementioned, we need to capture the relevant information with the input post and thus we need to predict the mediators in dialogue. Here, we propose to leverage Seq2seq-model with attention mechanism as the mediator predictor to generate {\it direct responding semantics}. This deterministic model can easily capture this simple semantic responding pattern and produce relevant response for our further processing.

\subsection{Auto-Regressive Response Generator}

\begin{figure*}[!t]
\centering
\vspace{-0.6cm}
\includegraphics[scale=0.35]{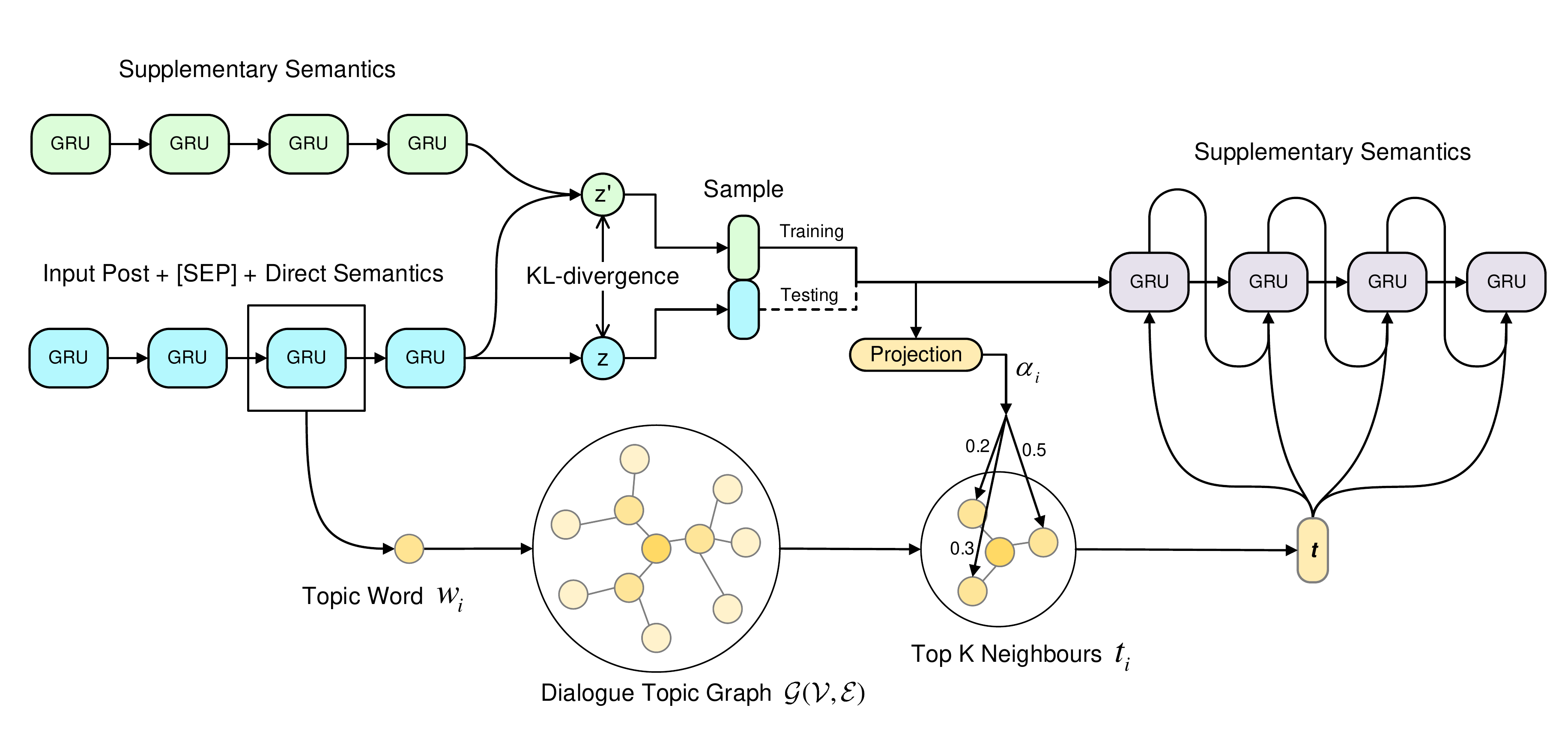}
\caption{The architecture of our proposed TGG-CVAE model.}
\label{fig2}
\vspace{-0.2cm}
\end{figure*}

So far, we have utilized the direct responding semantics generator to attain the mediator. Except for the path via mediator, the input post is also directly predictive of the diversified response. Now, based on the combination of input post and {\it direct responding semantics}, we aim to learn the {\it supplementary semantics}. The {\it supplementary semantics} is of great importance to provide useful information for interlocutors, and it can be rendered as an explanation, supplement, or extension of the previous  {\it direct responding semantics.} This semantic part has great diversity and contains many relevant entities. Although the high-quality observations of the mediators can reduce the confounding bias hidden in CVAE, it is not enough to attain one relevant and diversified {\it supplementary semantics} in the complex dialogue generation task. Following the previous causal analysis, we propose to exploit the dialogue topic graph to complement the semantic space and assign more relevant information into CVAE. Therefore, we design a novel model, {\it i.e.,} topic graph guided CVAE model (TGG-CVAE), to extend the semantic space in the conversation and sample more diversified and relevant sentences, and leverage the dynamic guidance from the dialogue topic graph to provide smooth and natural transition from the {\it direct responding semantics} to the {\it supplementary semantics}. The model structure is depicted in Figure \ref{fig2}.


To generate the supplementary semantics, the proposed TGG-CVAE model takes in the input post and previously generated direct semantics response. We denote the input $\vec{\oldhat{x}}$ as:
\begin{equation}
\vec{\oldhat{x}} = \{x_1, x_2, ..., x_T, [SEP], y_1, y_2, ..., y_{T^{'}} \},
\end{equation}
where the [SEP] token is a special token to separate the two sentences \cite{devlin2018bert}. The goal of this model is to generate the {\it supplementary semantics}:
\begin{equation}
  \vec{y}_{sup}=\{y_{T^{'}+1},y_{T^{'}+2},\cdots, y_{T^{''}}\}.
\end{equation}

This model mainly consists of four components: a prior network, a posterior network, a topic guide network, and a decoding network. The prior network is trained to approximate $p_\theta(z|\hat{x})$ while the posterior network is trained to approximate $q_\psi(z|\hat{x},\vec{y}_{sup})$, where $\theta$ and $\psi$ are the network parameters and $z$ is the latent variable. Here, $z$ is assumed to follow multivariate Gaussian distribution \cite{zhao2017learning} and then we have:
\begin{align}
 p_{\theta}(z|\hat{x}) & \sim \mathcal{N}(\mu, \sigma^2\mathbf{I}) \\
 q_\psi(z|\hat{x},\vec{y}_{sup}) &  \sim \mathcal{N}(\mu^\prime, \sigma^{\prime 2}\mathbf{I})
\end{align}
Typically, the prior network and the posterior network are RNN-based encoders that transform the input $\vec{\oldhat{x}}$ and $\vec{y}_{sup}$ into hidden states:
\begin{equation}
  \vec{h}^i_x  = f(\vec{h}^{i-1}_x,\vec{\oldhat{x}}_i)
\end{equation}
\begin{equation}
  \vec{h}^j_y  = f(\vec{h}^{j-1}_y,\vec{y}_j),
\end{equation}
where $i \in [1,T+T^{'}+1]$ and $j \in [T^{'}+1,T^{''}]$. The last hidden states from the prior/posterior network are denoted as $\vec{h}_x$ and $\vec{h}_y$ respectively. The latent variable is estimated by parameterizing its mean and log variance:
\begin{align}
    \begin{bmatrix}
        \mu \\
        \log(\sigma^2)
    \end{bmatrix} & = \vec{W}_p(\vec{h}_x) + b_p  \\
    \begin{bmatrix}
        \mu^\prime \\
        \log(\sigma^{\prime 2})
    \end{bmatrix} & = \vec{W}_r
    \begin{bmatrix}
        \vec{h}_x\\
        \vec{h}_y
    \end{bmatrix} + b_r,
\end{align}
where the $\vec{W}_p$, $\vec{W}_r$ and $b_p$, $b_r$ are the weights and biases for the prior network and posterior  network respectively. Reparametrization trick \cite{kingma2013auto} is used to keep the gradient propagate successfully in networks via a differentiable transformation of an auxiliary noise variable $\epsilon$:
\begin{align}
  z  &= \mu + \sigma\epsilon \\
  z^{\prime}  &= \mu^{\prime} + \sigma^{\prime}\epsilon
\end{align}

Then we can sample the latent variables $z$ or $z^{\prime}$ from the prior network or the posterior network. However, in the testing stage, as the ground-truth response is not available, the posterior latent variable $\vec{z^{'}}$ cannot be properly estimated. Therefore, we need to make sure that the prior network can fully acquire useful information from the posterior network by homogenizing $\vec{z}$ and $\vec{z^{'}}$. Here, KL-divergence loss is leveraged in our model to minimize the discrepancy between the two latent distributions:
\begin{align}
  \mathcal{L}_{KL}  =& KL(q_{\psi}(\vec{z}|\hat{x},\vec{y}_{sup})||p_\theta(\vec{z}|\hat{x})) \notag \\
                    =& \int q_{\psi}(\vec{z}|\hat{x},\vec{y}_{sup}) log \frac{q_{\psi}(\vec{z}|\hat{x},\vec{y}_{sup})}{p_\theta(\vec{z}|\hat{x})} dz,
\end{align}
from which we can derive the final formula for calculating KL-divergence loss:

\begin{equation}
  \mathcal{L}_{KL} = log\frac{\sigma}{\sigma^{\prime}} + \frac{\sigma^{\prime2}+(\mu-\mu^{\prime})^2}{2\sigma^2} - \frac{1}{2}
\label{eqa:cvae}
\end{equation}


For the vanilla CVAE model, $\vec{z}$ or $\vec{z^{'}}$ is directly fed as the input of the decoder for decoding from the semantic space. However, as aforementioned, the latent semantic space is too large to train well and the sampling steps in prior and posterior distributions of vanilla CVAE will act as the confounder between the input combination (input post and {\it direct responding semantics}) and {\it supplementary semantics}. Because the input post and response pairs in the real data maybe do not conform to the assumed prior or posterior distributions of CVAE, this confounding bias may make the model learn the spurious statistical cues for prediction of diversified response, resulting in some linguistically similar but inconsistent or irrelevant expressions in the generated sentences. Therefore, reducing the confounding bias is essential for the {\it supplementary semantics} generation. We exploit the dialogue topic graph to complement the semantic space and assign more accurate and relevant relationship into CVAE so as to mitigate the confounding bias. Details of this strategy are as follows:

Firstly, the topic words $w_1, w_2, ..., w_m$ are extracted from $\vec{\oldhat{x}}$ using the TF-IDF method, and then they are placed into the dialogue topic graph $\mathcal{G(V,E)}$ to find their nearest $n$ neighbours $w_{11}, w_{12}, ..., w_{mn}$ according to the weight, where $w_{ij}$ is the $j$-th neighbours of the topic word $w_{i}$. We then choose the neighbour with the highest probabilities:
\begin{equation}
  t_1, ...,t_K = \mathop{{\rm arg max}_K}_{i\in [1,m],j\in [1, n]}(\Pr(w_{ij}|w_i)),
\end{equation}
to select top K topic words, namely, $t_1, ...,t_K$.

Secondly, since these topic words contribute differently to the generation of a response, we leverage the sampled latent variables to formulate a dynamic prior/posterior selection of the topic words. The sampled latent variables  $\vec{z}$ (testing) or $\vec{z^{'}}$ (training) are passed through a projection layer to produce a distribution over the K topic words, namely $\vec{\alpha} =\alpha_1,\alpha_2,...,\alpha_K $. The final representation of the topic information is formulated as a weighted summation of the topic embeddings:
\begin{equation}
  \vec{t} = \alpha_i \cdot \vec{t_i}, \quad i=1,2,3,...,k
\end{equation}
where $\vec{t_i}$ is the embedding of the word $t_i$.

Thirdly, the topic information $\vec{t}$ and the sampled latent variable $\vec{z}$ or $\vec{z^{'}}$ are fed into the decoder for generating the supplementary semantics:
\begin{equation}
\Pr ( y_t|y_{1:t-1}, \vec{x}, \vec{y}_{direct})=g(\vec{y}_{t-1},\vec{s_t},\vec{z},\vec{t}),
\end{equation}
from which we can find that each supplementary semantic word is conditioned on both the topic information and the sampled latent variable, and thus the sentences can be related to the previous words and have more diversity. Following \cite{sohn2015learning}, we train the proposed model by maximizing the variational lower bound of the conditional log likelihood:
\begin{align}
  \mathcal{L}_{ELBO} =& -KL(q_{\psi}(\vec{z}|\hat{x},\vec{y}_{sup})||p_\theta(\vec{z}|\hat{x})) \notag \\
  & + {\rm E}_{q_{\psi}(\vec{z}|\hat{x},\vec{y}_{sup})}[{\rm log}\, p_\theta(\vec{y}_{sup}|\vec{z},\hat{x},\vec{t})],
\end{align}
where the $KL(.,.)$ denotes the KL-divergence of two distributions. Since the latent semantic distribution is easy to collapse (\aka  KL-collapse problem), we add a bag-of-words loss $\mathcal{L}_{BOW}$ and use KL-annealing strategy to deal with this problem \cite{zhao2017learning}. The final loss function of this proposed model is formulated as:
\begin{equation}
  \mathcal{L} = \mathcal{L}_{ELBO} + \mathcal{L}_{BOW} + \mathcal{L}_{direct}
\end{equation}

Note that, we also consider the possible circumstance where the responses do not contain any {\it supplementary semantics} by leveraging the [EOS] token as the placeholder. If the TGG-CVAE model predicts [EOS] token at the first step, this indicates that the {\it direct responding semantics} is already complete and it does not need any {\it supplementary semantics}.

\section{Experimental Results}
\vspace{-2mm}
\label{sec:exp}
\subsection{Dataset}
\vspace{-2mm}

We conduct experiments on a large-scale real-world dialogue dataset, {\it i.e.,} Short-Text Conversation (STC) dataset \cite{shang2015neural}. This dataset is publicly available and is cleaned by the data publishers. It consists of 4,433,853 post-comment pairs collected from Chinese Weibo, a social media platform where people can chat online. 




\subsection{Evaluation Metrics}
\vspace{-2mm}
\begin{table*}[!t]
\vspace{-0.5cm}
\centering
\begin{tabular}{l|c|c|c|c|c|c}
\textbf{Models}      & \textbf{BLEU-1} & \textbf{BLEU-2} & \textbf{BLEU-3} & \textbf{BLEU-4} & \textbf{Distinct-1} & \textbf{Distinct-2} \\
\hline
\textbf{Seq2seq} \cite{sutskever2014sequence}  & 0.2392           & 0.1937            & 0.1646          & 0.1304           & 0.0549          & 0.1859     \\
\textbf{CVAE} \cite{zhao2017learning}          & 0.2223           & 0.1808            & 0.1541          & 0.1222           & 0.0936          & \underline{0.4208}    \\
\textbf{MMPMS} \cite{chen2019generating}       & 0.2246           & 0.1868            & 0.1612          & 0.1289           & \textbf{0.0972} & \textbf{0.4214}     \\
\textbf{DCVAE} \cite{gao2019discrete}          & 0.2124           & 0.1700            & 0.1436          & 0.1134           & 0.0405          & 0.1681     \\ \hline
\textbf{SLARM} (ours)                                & \textbf{0.2657}  & \textbf{0.2169}   & \textbf{0.1850} & \textbf{0.1469}  & 0.0879          & 0.3685     \\
\textbf{SLARM} w/o TGG    (ours)                     & \underline{0.2569}           & \underline{0.2099}            & \underline{0.1792}          & \underline{0.1423}           & \underline{0.0967}          & 0.4088    \\
\textbf{SLARM} w/o CVAE    (ours)                    & 0.2544           & 0.2069            & 0.1763          & 0.1398           & 0.0881          & 0.2195     \\
\end{tabular}

\caption{Automatic evaluation results on STC dataset. The best results are in boldface and the second best results are underlined.}
\label{bleu:result}
\end{table*}

\begin{table*}[!t]
\centering
\begin{tabular}{l|c|c|c|c}
\textbf{Models}      & \textbf{Relevance} & \textbf{Informativeness} & \textbf{Fluency} & \textbf{Average} \\
\hline
\textbf{Seq2seq} \cite{sutskever2014sequence} & 1.52           & 1.63           & \textbf{2.68} & 1.94         \\
\textbf{CVAE} \cite{zhao2017learning}         & 1.45           & 1.73           & 2.49          & 1.89            \\
\textbf{MMPMS} \cite{chen2019generating}      & 1.54           & \textbf{2.02}  & 2.00          & 1.85           \\
\textbf{DCVAE} \cite{gao2019discrete}         & \textbf{1.96}  & 1.53           & 2.48          & \underline{1.99}       \\ \hline
\textbf{SLARM}  (ours)                        & \underline{1.57}           & \underline{1.82}           & \underline{2.67}          & \textbf{2.02}   \\
\end{tabular}
\caption{Human evaluation results on STC dataset. The best results are in boldface and the second best results are underlined.}
\label{human:result}
\vspace{-0.5cm}
\end{table*}

\paratitle{Automatic Evaluation}. We adopted two widely-used metrics, {\it BLEU-n} \cite{papineni2002bleu} and {\it Distinct-n}  \cite{li2015diversity}, to automatically evaluate the dialogue generation models. {\it BLEU-n} score is a referenced evaluation metric to measure word overlap between the generated response and the reference.
%
%
Note that in our experiment we apply smoothing function 7 \cite{chen2014systematic}  to avoid the problem when no $n$-gram overlaps are found. {\it Distinct-n} score \cite{li2015diversity} is used to determine word-level diversity of the generated response. It is measured by calculating the percentage of distinct $n$-grams in the generated responses.

\paratitle{Human Evaluation}. We randomly sampled 100 posts from the test set and let the models generate corresponding responses. Three annotators were invited to rate the post-response pairs from three aspects: relevance (whether the response is relevant to the input post), informativeness (whether the response is informative) and fluency (whether the response has no grammar mistakes).
A three-point scale (0,1,2) is used in the evaluation for the above aspects. When contradiction occurs between the first two annotators, the third annotator will resolve the disagreement. Fleiss’ kappa \cite{fleiss1973equivalence} is calculated to measure the inter-rater agreement between the first two annotators.

\subsection{Baseline Models}
\textbf{Seq2seq} \cite{bahdanau2014neural}: it is a canonical seq2seq model with the attention mechanism.
\textbf{CVAE} \cite{zhao2017learning}: it is a conditional variational auto-encoder model. During testing, we randomly sample latent variables from the prior network and generate corresponding responses.
\textbf{MMPMS} \cite{chen2019generating}: it is a multi-mapping and posterior mapping selection model. We use their original implementation and hyper-parameter settings. 
\textbf{DCVAE} \cite{gao2019discrete}: it is a discrete CVAE model. We use their original implementation and adopt the two-stage sampling strategy during testing.

\subsection{Implementation Details}
For our approach, we use 2-layers GRU units for encoders in the prior network/posterior network and the hidden size is set to 256. The embedding size and vocabulary size are set to 200 and 40,000 respectively. Word embeddings are randomly initialized and OOV (out-of-vocabulary) words are replaced with a special token UNK.  Adam optimizer \cite{kingma2014adam} is used for optimization and the training batch size is 128. The initial learning rate is set to 0.5 and a learning rate decay operation is employed when the validation loss stops decreasing for three consecutive epochs. The decay rate is 0.99. The top 5 neighbors of the topic words in the dialogue graph are chosen and fed into the decoder.

\subsection{Results}

\paratitle{Automatic evaluation} results are shown in Table \ref{bleu:result}. Notably, our SLARM model outperforms all of the baselines in terms of BLUE metric (with p-value $<$ 0.05) and its performance is 11.2\% ahead of the second best model. This verifies our assumption that splitting the to-be-generated responses into different semantic parts and separately generating them with suitable methods will enhance the overall performance. As for the Distinct metric, the performance of our model is moderate compared to the CVAE model and MMPMS model. This is because our main objective is not only boosting the diversity of responses but also promoting relevance between posts and generated responses.

To further analyze the results, we conduct ablation studies by removing the Topic Graph Guided module (\ie SLARM w/o TGG) or replacing the CVAE module with traditional GRUs  (\ie SLARM w/o CVAE). After removing TGG, the Distinct performance increases and the BLEU performance decreases. This indicates that our dynamic topic graph guiding strategy is effective in providing relevant information from posts and thus can increase BLEU scores. However, this strategy gets slightly lower Distinct scores because the restrained topics would reduce possibilities in choosing more diversified words. When the CVAE module is removed, the Distinct-2 score drops by a large margin, indicating the CVAE module is effective in extending the semantic space and sampling diversified phrases. The BLUE scores also decrease because the posterior network is essential in providing additional information to dynamically weigh the contribution of topic words. Hence, each component of our model complements each other, and thus the model has the self-adaptive capability to reach a balance between diversity and relevance.

\paratitle{Human evaluation} results are shown in Table \ref{human:result}. The DCVAE model surpasses other models in relevance metric. This is owing to the fact that DCVAE model tends to re-use the words in the posts to generate a response, which makes the annotators give high relevance scores. The discrete latent variables from the prior and posterior network are pre-trained to predict keywords in the post, and thus sampling from these variables tends to produce the same words in the post. However, the latent variables constrain the generation process, which leads to low informative scores. The MMPMS model performs the best from the informativeness aspect. This is because the auxiliary loss (\ie matching loss) is effective in encouraging the selection module to choose different and diverse mapping modules. However, some mapping modules are not well-trained and they generate ungrammatical sentences. Therefore their fluency score is rather low. The Seq2seq model gets the highest fluency score, as it often generates common and simple sentences.

Our proposed SLARM model outperforms all the baseline models in terms of the average score. For every single aspect, the SLARM model consistently obtains the second best scores. The second best relevance score indicates that first generating the {\it direct responding semantics} will assure the relevance with the post because it directly answers the question. The second best informative score shows that the proposed SLARM model can enhance the diversity and generate informative sentences. Our fluency score is also the second best and is close to the Seq2seq model's, which verifies that our methods can alleviate the grammatical problems when concatenating two semantic parts.

Note that the Fleiss’ kappa for relevance, informativeness, and fluency are 0.4153, 0.4188, and 0.4378, respectively, indicating ``moderate agreement'' among the annotators.

\subsection{Case Study}

\begin{figure*}[h!]
\centering
\includegraphics[scale=0.5]{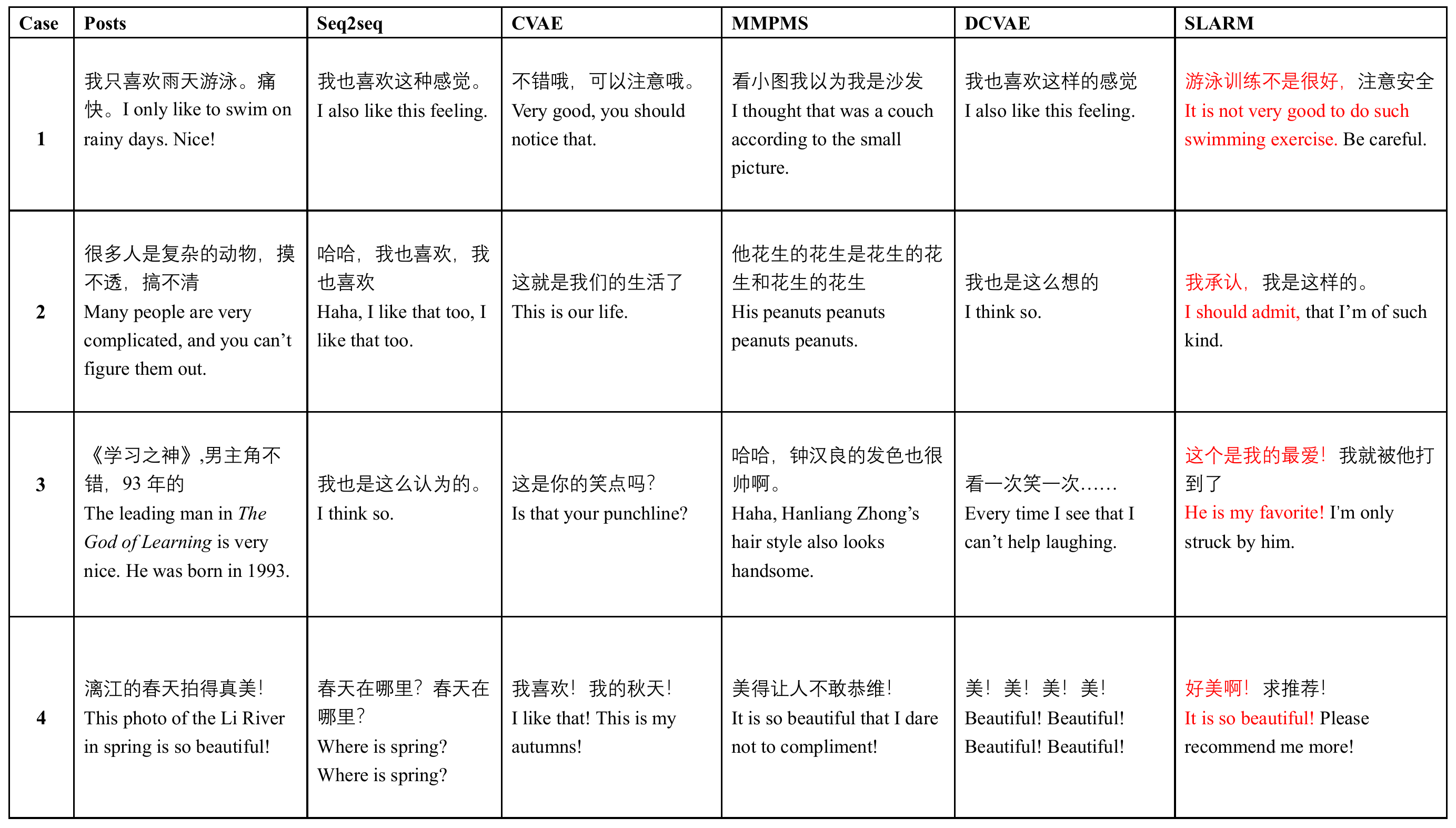}
\vspace{-3mm}
\caption{Case study of the sampled 4 cases. For the SLARM model, words in red are generated by the {\it direct responding semantics} generator, while the rest are generated by  {\it supplementary semantics} generator.}
\vspace{-3mm}
\label{casestudy}
\end{figure*}

We present sampled 4 cased in our Appendix. As is shown in the figure, the Seq2seq model tends to generate safe and generic responses, such as case 1, 2, and 3. The response pattern generated by Seq2seq models often starts with ``I also like...'' or ``Haha...'', which makes the responses dull and boring. However, in cases 1 and 3, although these responses are generic, they are semantically appropriate and relevant according to the post. Therefore, this model is suitable for searching for the mediators in dialogue generation, which is actually observed in our model by {\it direct responding semantics}.
The CVAE model is better at choosing diversified words such as ``punchline'' in case 3 and ``autumn'' in case 4. However, the confounding bias makes this model learn some spurious statistical cues for predicting diversified words, and thus these words are not logical with regard to the input question.
The MMPMS model can produce informative sentences, such as in cases 1, 3, and 4. In case 3, the MMPMS model produces a response that is not only informative but also relevant to the input post, but responses in case 1  are irrelevant. Besides, another major problem is that some of the mapping modules are not well-trained and thus in case 2 we can see the generated sentence is ungrammatical and irrelevant.
The DCVAE model tends to copy the input post, such as in case 4. This is the reason why the relevant score for DCVAE model is higher than other models. In some circumstances, DCVAE produces the same results as Seq2seq, such as in case 1 and 2. 

We can conclude that the SLARM model performs the best and reaches a balance between relevance and diversity. The {\it direct responding semantics} (marked in red) in case 1, 2, 3, and 4 are very relevant to the input post, and the {\it supplementary semantics} provide more and diversified information to complete the response. In case 1, the {\it supplementary semantics} is generated to provide further instruction of being careful when swimming on rainy days. In case 3, the {\it supplementary semantics} re-emphasizes that the interlocutor is fond of the actor. Additionally, we can observe from the cases that with the dialogue topic graph guiding strategy, the transition from {\it direct responding semantics} to {\it supplementary semantics} is smooth and natural. Therefore, these cases fully demonstrate the model's capacity for generating the relevant and diversified responses via searching for the direct responding semantic parts as mediators in dialogues and then utilizing our proposed SLARM model to mitigate the confounding bias and thus enhance the diversity without the loss of relevance.

\section{Conclusion}
\vspace{-2mm}
\label{sec:con}
In this paper, we incorporate the causal analysis into the dialogue generation task by searching for the mediators and mitigating the confounding bias in dialogues. We thus propose a sentence level auto-regressive response generation model to first generate mediators to preserve relevance with the input post, and then generate the diversified semantics based on our proposed (SLARM) model. Extensive experimental results demonstrate the effectiveness of our approach. For future work, we are exploring more complicated and self-adaptive methods for locating mediators, and we are trying to leverage de-confounding methods to deal with the CVAE problem.

\section{Acknowledgments}
This work was supported in part by the National Natural Science Foundation of China under Grant No.61602197, Grant No.L1924068, Grant No.61772076, Grant No.62276110, in part by CCF-AFSG Research Fund under Grant No.RF20210005, and in part by the fund of Joint Laboratory of HUST and Pingan Property \& Casualty Research (HPL).

\bibliography{acl_latex}
\bibliographystyle{acl_natbib}


\end{document}